\documentclass[letterpaper, 10 pt, conference]{ieeeconf}  

\IEEEoverridecommandlockouts                              

\usepackage{graphics} 
\usepackage{epsfig} 
\usepackage{amsmath} 
\usepackage{amssymb}  
\usepackage{pifont}  
\usepackage{xcolor}  
\usepackage[section]{placeins} 
\usepackage{siunitx}
\usepackage{multirow}

\usepackage{url}

\newcommand{\cmark}{\textcolor{green!60!black}{\ding{51}}}
\newcommand{\xmark}{\textcolor{red!80!black}{\ding{55}}}

\title{\LARGE \bf
ViHaTeleop: A Low-Cost, Lightweight Visual-Haptic Teleoperation System for Dexterous Manipulation Learning
}

\author{Fucai Zhu$^{1}$, Yanhou Lai$^{1}$, Paul Maestre$^{1}$, and Koichi Hashimoto$^{1}$%
\thanks{$^{1}$Graduate School of Information Sciences, Tohoku University, Sendai, Japan. {\tt\small zhu.fucai.s6@dc.tohoku.ac.jp}. Project website: \protect\url{https://laiyanhou.github.io/ViHaTeleop-website/}}%
}
\begin{document}

\maketitle
\thispagestyle{empty}
\pagestyle{empty}

\begin{abstract}
Learning from demonstration is a promising approach for dexterous manipulation, but collecting high-quality contact-critical demonstrations remains difficult with low-cost teleoperation hardware. We present ViHaTeleop, a lightweight (0.7 kg), low-cost (\$550) visual-haptic teleoperation system with SLAM-based wrist tracking, camera-based hand tracking, and finger-wise vibrotactile feedback through Linear Resonant Actuators (LRA). The system includes several design choices (LED illumination, fisheye hand camera, and tactile-aware retargeting constraints) and is deployed on Franka + LEAP Hand + 9DTact in both real and simulated environments. Under matched with/without-haptic conditions with nine participants across six contact-critical tasks, haptics improved success rates across all tasks (+2.2 to +15.6 percentage points), while completion-time effects were task-dependent. Subjective ratings showed significant gains in contact clarity and grasp confidence in both simulation and real-world settings (Wilcoxon signed-rank, $p<0.05$). We also integrate a lightweight depth-camera-based tactile proxy in Isaac Sim, enabling a full pipeline from multi-modal demonstration collection to visual-tactile policy training. Preliminary downstream validation by training visual-tactile policies from collected demonstrations shows tactile cues benefit contact-critical subtasks (peg-in-hole: +17 percentage points over vision-only).

\end{abstract}

\section{INTRODUCTION}
\begin{figure}[t]
    \centering
    \includegraphics[width=\columnwidth]{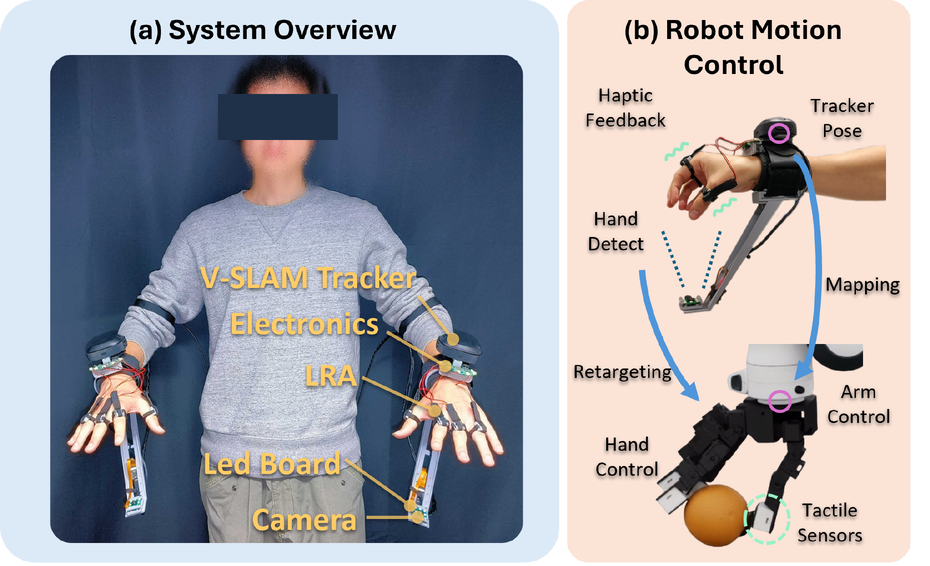}
    \caption
    {Hardware overview and control architecture of ViHaTeleop. (a) The system consists of two bimanual wrist and hand configuration tracking modules. Each module uses a VSLAM+IMU tracker (HTC Vive Ultimate) for wrist pose tracking and a fisheye camera for hand configuration detection in real-time. LRA rings on the operator's thumb, index, and middle fingers provide vibrational haptic feedback. (b) The captured wrist and hand configurations are mapped and retargeted to control a robotic arm (Franka) with a dexterous hand (LEAP Hand). Tactile sensors (9DTact) on the robot fingertips provide touch sensory feedback to the operator.}
    \label{fig:sys_overview}
\vspace{-8mm}
\end{figure}
Large-scale robot manipulation data is critical for embodied AI, with teleoperation-based data collection playing an increasingly important role \cite{openx_icra24, droid_arxiv24, agibot_iros25, bridgedata_rss22, bridgedata_v2}. However, current teleoperation systems face a fundamental limitation: expensive systems ($>$\$10k) provide haptic feedback for contact-critical manipulation, while affordable systems ($<$\$1k) lack haptic feedback entirely. This gap particularly hinders data collection for fine-grained manipulation tasks requiring delicate force control—such as grasping fragile objects (e.g., potato chips) or performing precise assembly—where haptic feedback is essential for operator awareness and performance.

In this paper, we define contact-critical tasks as tasks where (i) success depends on detecting contact onset and maintaining stable contact, and (ii) errors are strongly coupled to interaction forces (e.g., slip, breakage, jamming, or insertion failure). Visual constraints are treated as an evaluation context rather than part of the definition.

Our primary research question is whether low-cost vibrotactile feedback measurably improves contact-critical teleoperation quality in a matched vision-based setup. While several systems incorporate haptic feedback \cite{nuexo, bunnyvp}, the impact of low-cost vibrotactile actuators has not been systematically quantified under matched with/without-haptic conditions. In our 9-participant study, haptic feedback produced descriptive success-rate gains across all tasks, while completion-time effects were task-dependent. As a secondary objective, we evaluate whether demonstrations collected with this system are usable for downstream visual-tactile imitation learning.

ViHaTeleop (illustrated in Fig.~\ref{fig:sys_overview}) consists of an off-the-shelf VSLAM+IMU-based tracker (HTC Vive Ultimate \cite{vive_product}) for capturing wrist motion, a fisheye camera (160-degree field of view, \$11) for detecting hand configuration, and LRA rings (\$1.4 each) for generating vibrational haptic feedback. The camera is wrist-mounted, moving together with the hand—similar to ACE \cite{ace}—but with significantly reduced mass and torque. During operation, we capture the tracker pose and hand configuration in real-time. The tracker pose is mapped to the robot arm end-effector pose to control the Franka \cite{franka} arm via inverse kinematics with joint position commands. The hand configuration is retargeted to the LEAP Hand\cite{leaphand} joints using the method from \cite{anyteleop}. The 9DTact \cite{9dtact} sensors (a low-cost, open-source tactile sensor) mounted on the LEAP Hand fingertips capture real-time tactile signals, which are converted to PWM control signals for the LRAs via a driver board mounted on the operator's wrist.  We implement ViHaTeleop to teleoperate a Franka + LEAP Hand + 9DTact platform both in reality and simulator. In simulation, we additionally provide a lightweight 9DTact approximation in Isaac Sim 5.0 \cite{isaacsim} using a depth-camera–based proxy.

\begin{table}
\centering
\caption{Comparison of Teleoperation Systems}
\label{tab:comparison}
\footnotesize
\setlength{\tabcolsep}{3pt}
\begin{tabular}{lcccc}
\hline
\textbf{System} & \textbf{Cost} & \textbf{Wearable} & \textbf{Haptic} & \textbf{Occlusion} \\
& \textbf{(USD)} & \textbf{Burden (kg)} & \textbf{Feedback} & \textbf{Robust} \\
\hline
NuExo \cite{nuexo} & $>$10k & Heavy (5.2) & \cmark & \cmark \\
BunnyVP \cite{bunnyvp} & 4k & Light (0.65)$^*$ & \cmark & \xmark \\
DexCap \cite{dexcap} & 3.6k & Light (1.8)$^\dagger$ & \xmark & \cmark \\
AnyTeleop \cite{anyteleop} & 50 & None & \xmark & \xmark \\
ACE \cite{ace} & 600 & Medium & \xmark & \cmark \\
\hline
\textbf{Ours} & \textbf{550} & \textbf{Light (0.7)} & \textbf{\cmark} & \textbf{\cmark} \\
\hline
\multicolumn{5}{l}{\scriptsize $^*$Vision Pro headset weight. $^\dagger$Full backpack system; glove portion is lighter.}
\end{tabular}
\vspace{-8mm}
\end{table}

As shown in Table~\ref{tab:comparison}, ViHaTeleop achieves a favorable balance across cost, weight, haptic feedback, and occlusion robustness compared to recently proposed systems. Through targeted design improvements (LED lighting for varied illumination, a fisheye camera to reduce mounting lever arm and torque, intuitive tactile feedback, and retargeting constraints for flat-surface tactile sensors), ViHaTeleop provides a practical operating point for contact-critical data collection.

While ViHaTeleop builds upon established teleoperation techniques, we summarize our key contributions as follows:
\begin{enumerate}
\item Primary: controlled evaluation of low-cost vibrotactile feedback under matched with/without-haptic conditions across contact-critical tasks (user study).
\item A complete haptic integration pipeline: tactile-aware retargeting constraints (enabling multi-modal data quality), multi-fingertip 9DTact sensing, and LRA vibrotactile feedback mapping, on a low-cost (\$550), lightweight (0.7 kg) platform.
\item A simulation-integrated pipeline covering the full loop from multi-modal data collection to policy evaluation: Isaac Sim task environments, a lightweight depth-camera-based 9DTact proxy, and multi-fingertip tactile point cloud registration for the LEAP Hand. Preliminary validation using an extended 3D-ViTac architecture (first applied to $>$20-DoF dexterous manipulation) demonstrates usable visual-tactile demonstrations for downstream imitation learning.
\end{enumerate}
The scope of this paper is primarily the teleoperation system study; the imitation-learning section is included as a constrained downstream feasibility validation rather than a standalone algorithmic contribution.

\section{Related Work}

\subsection{Dexterous Teleoperation}

\textbf{Hand Teleoperation:} Dexterous hand teleoperation encompasses diverse approaches, each presenting distinct tradeoffs. Vision-based methods using a single camera only \cite{anyteleop,dexpilot}  or  VR headsets \cite{vpteleop,bunnyvp} are low-cost and lightweight but struggle with occlusion issues. Optical tracking systems \cite{VATO}, while more accurate, remain costly, non-portable, and still susceptible to occlusion despite multi-camera setups.
Wearable sensor approaches \cite{dexcap}, Manus®, Dexmo®, SenseGlove®—circumvent occlusion entirely by directly measuring joint angles. However, they are expensive and difficult to add customization for other feedback. Custom exoskeletons \cite{vitacformer,dexumi,dexop} physically replicate robot kinematics, enabling intuitive control at the expense of substantial design effort, limited cross-platform generalizability, and considerable operator burden—particularly when human and robot morphologies differ.
With the anti-occlusion improvement from \cite{ace}, and hand detecting algorithms \cite{frankmocap,mediapipe} and efficient retargeting algorithm \cite{dexpilot,anyteleop,bunnyvp} mapping human hand configuration to various anthropomorphic robot hands (LEAP Hand\cite{leaphand}, etc), the vision-based method works as a promising method for teleoperating dexterous robot hands to meet mass data collection purposes.
DexPilot \cite{dexpilot} introduced distance-dependent projection mechanisms for precise fingertip control, which we adopt as our retargeting foundation. We extend this framework with tactile-aware constraints tailored for flat-surface sensors (Sec.~\ref{sec:retargeting}).

\textbf{Wrist Teleoperation:} Fixed kinematic linkages \cite{aloha,gello} provide stable and accurate tracking but lack portability. Wearable exoskeletons \cite{airexo,airexo2,ace} improve portability but introduce considerably greater weight compared to tracker-based alternatives. Recent evaluation of SLAM-based tracker HTC Vive Ultimate \cite{vivetracker} demonstrates performance competitive with or superior to low-cost kinematic systems using 3D-printed components and encoders, while maintaining minimal operator burden and high portability.

\textbf{Teleoperation with Haptic Feedback:} Integrating haptic feedback to improve teleoperation and the sense of tele-existence is a long-standing research area. Various mechanisms have been explored. Vision-only proxies for touch \cite{rdp,3dvitac} can help collect delicate manipulation data; however, extended reliance on visual perception alone is cognitively demanding and less intuitive for operators. Motorized force-feedback devices \cite{doglove,hexotrac,vitacformer}—including off-the-shelf options such as Dexmo® and SenseGlove®—provide richer cues but are too expensive and too heavy. Electrotactile methods \cite{electrohaptic1,electrohaptic2} offer high spatial resolution, yet their fabrication processes are not mature enough for mass production. By contrast, low-cost, robust vibrotactile feedback \cite{bunnyvp} can improve performance without increasing operator burden. Moreover, a proper ring-mounted design is compatible with vision-based methods. We further conduct a user study to validate the haptic feedback integrated into our system, using thorough objective and subjective metrics.

While systems like NuExo \cite{nuexo} and ACE \cite{ace} provide valuable capabilities, each presents distinct limitations for large-scale data collection: NuExo's weight (5.2 kg) and cost ($>$\$10k) hinder prolonged use and mass deployment, while ACE's exoskeleton adds considerable weight, and its mobile version suffers from unstable end-effector mapping due to a non-tracked floating base. We adopt ACE's hand-tracking approach but replace the exoskeleton with SLAM-based tracking, providing a fixed-origin reference frame for more intuitive and stable mapping. Importantly, ViHaTeleop maintains interface consistency with ACE \cite{ace}, enabling potential extension to other robotic platforms beyond Franka + LEAP Hand.

\subsection{Visual-Tactile Policy Learning from Human Demonstrations}
Though still under-explored, visual-tactile policy learning from human demonstrations is gaining increasing attention, with several recent works integrating tactile modality alongside vision \cite{rdp,3dvitac,vitacformer,VATO}. When combined with high-DoF dexterous hands \cite{leaphand} and high-resolution tactile sensors \cite{gelsight,9dtact}, policy learning becomes a challenging high-dimensional problem. The imitation learning method 3D-ViTac \cite{3dvitac} offers a promising approach by integrating visual and dense-array tactile data into a unified 3D point cloud representation, demonstrating strong potential for dexterous hand manipulation. Given the high-quality, contact-critical, long-horizon manipulation data collected by our teleoperation system, we extend 3D-ViTac as a downstream learning module to evaluate our complete pipeline from data collection to policy execution.

\section{System Design}
The primary design targets are low cost ($<$\$550 using off-the-shelf components), lightweight wearable (0.7\,kg via hollow 3D-printed linkages and fisheye camera), robustness (LED lighting for varied illumination), portability (modular snap-fit assembly in minutes), and haptic feedback integration---all essential for scalable contact-critical data collection.
\subsection{Hardware Design}
\begin{figure}[t]
    \centering
    \includegraphics[width=\columnwidth]{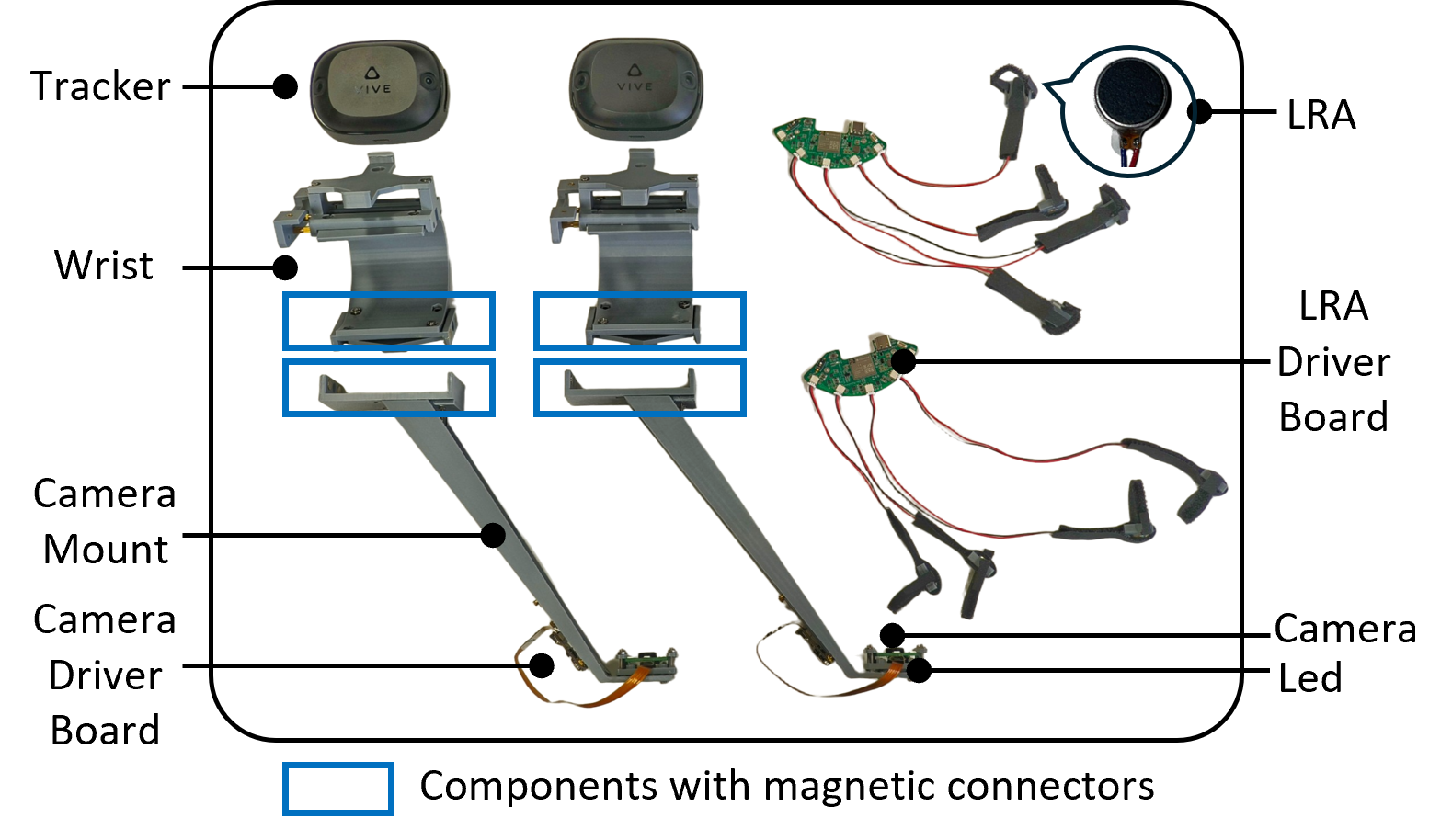}
\vspace{-8mm}
    \caption{Hardware Components.}
    \label{fig:hardware}
\vspace{-7mm}
\end{figure}

The hardware consists of four primary modules (Fig.~\ref{fig:hardware}): (1) \textit{Tracker module}: HTC Vive Ultimate tracker for 6-DoF wrist pose tracking via VSLAM and IMU, communicating wirelessly over Wi-Fi; (2) \textit{Camera module}: OV5647 with 160° fisheye lens for hand configuration detection, connected via USB; (3) \textit{LED module}: Ring-mounted LEDs powered through USB for illumination enhancement; (4) \textit{Haptic module}: three LRA rings worn on the operator's thumb, index, and middle fingers---matching the three 9DTact-instrumented fingertips---driven by a custom PCB with ESP32 microcontroller, connected via USB.
We designed a custom detachable connector for the bottom-left interface of the Vive Tracker, enabling quick, tool-free attachment to the 3D-printed mount. The connector ensures stable fixation and accurate pose alignment while allowing easy removal for charging or reconfiguration. Combined with magnetic snap-fit joints between printed parts, all ViHaTeleop submodules can be detached and reassembled within minutes, enhancing portability and user convenience.

\subsection{Pose Estimation}
\begin{figure}
    \centering
    \includegraphics[width=0.6\columnwidth]{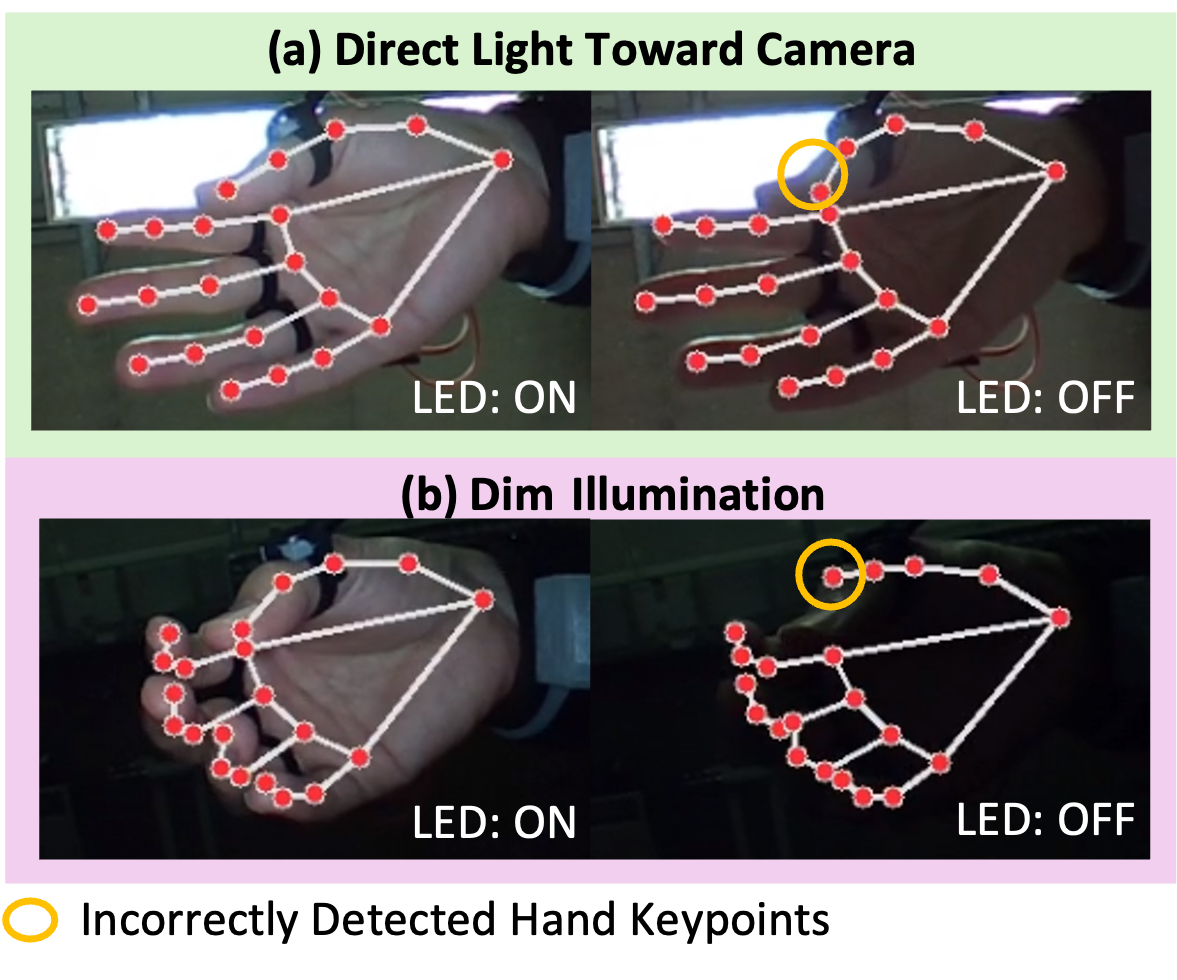}
\vspace{-5mm}
    \caption{Hand skeleton detection in challenging lighting: (a) direct glare toward camera; (b) low-light scene. LED illumination stabilizes detection.}
    \label{fig:illumination}
\vspace{-8mm}
\end{figure}

The operator's wrist pose is captured by the tracker via VSLAM and IMU sensing, providing 6-DoF tracking. Hand images from the wrist-mounted fisheye camera are undistorted using pre-calibrated intrinsic parameters and processed with MediaPipe \cite{mediapipe} to detect 21 hand keypoints in real-time.
Integrated LEDs support hand tracking under adverse illumination, with representative ON/OFF comparisons shown in Figure~\ref{fig:illumination}. In a separate single-participant evaluation comprising four gestures, four lighting/LED conditions, and ten repetitions each (160 trials), LEDs reduced mean frame-to-frame landmark jitter by 39.6\% under direct glare and 21.3\% in low light, while increasing low-light detection from 98.3\% to 100\%.

\subsection{Robot Arm Control}
The tracker provides both position and orientation. We map the tracker motion to a robot end-effector target pose using relative pose mapping: the operator's initial wrist pose defines a reference frame, and subsequent wrist motion is computed as a relative transform. To handle workspace mismatch, we clamp target translation to a predefined workspace box in the robot base frame. The target pose is executed through inverse kinematics with joint limits; if IK fails or violates limits, we hold the last valid command to avoid discontinuities. The initial robot configuration is set to a fixed nominal pose, and the initial tracker pose is used as the teleoperation reference.
\subsection{Robot Hand Motion Retargeting}
\label{sec:retargeting}
DexPilot retargets human keypoints to LEAP Hand joint angles via optimization, encouraging fingertip geometry matching while regularizing joint motion. A distance-dependent projection/weighting increases fingertip emphasis as the thumb approaches a finger to improve pinch precision. We smooth this schedule because abrupt thresholding can cause discontinuous intermediate grasp configurations (``snapping'').

\textbf{Tactile-Aware Parallelism Constraint:}
To maximize contact quality with flat-surface tactile sensors (e.g., 9DTact \cite{9dtact}, GelSight \cite{gelsight}), we extend DexPilot's framework by introducing a fingertip parallelism constraint. Standard retargeting optimizes for position matching but does not enforce orientation alignment, often resulting in edge contact that limits tactile sensing effectiveness. We augment the cost function with an alignment term:
\begin{equation}
\mathcal{L}_{\text{parallel}} = w_p \cdot \sum_{j \in \mathcal{F}} \|n_{\text{thumb}} \cdot n_j + 1\|^2
\end{equation}
where $\mathcal{F} = \{\text{index, middle, ring}\}$,  and $n_j$ denotes the outward surface normal of fingertip $j$ computed from auxiliary URDF links. To prevent abrupt "snapping" in standard DexPilot projection, we extend its projection mechanism with adaptive functions that smoothly interpolate based on thumb-to-finger distance $d$:
\begin{equation}
\begin{split}
\eta(d) &= \eta_{\min} + \frac{d - d_0}{d_{\max} - d_0}(\eta_{\max} - \eta_{\min}) \\
w(d) &= w_{\min} + \frac{d_{\max} - d}{d_{\max} - d_{\min}}(w_{\max} - w_{\min})
\end{split}
\end{equation}
where $\eta(d)$ controls projection distance and $w(d)$ modulates retargeting strength. For the LEAP Hand, we set $d_{\min} = d_0 = 30$mm, $d_{\max} = 50$mm, $\eta_{\min} = 0.1$mm, $\eta_{\max} = d_{\max}$, $w_{\min} = 1.0$, $w_{\max} = w_p = 200$. The parallelism weight $w_p$ activates when $d < d_{\max}$, enabling smooth fingertip approach with intermediate grasp configurations.
\begin{figure}
    \centering
    \includegraphics[width=0.85\columnwidth]{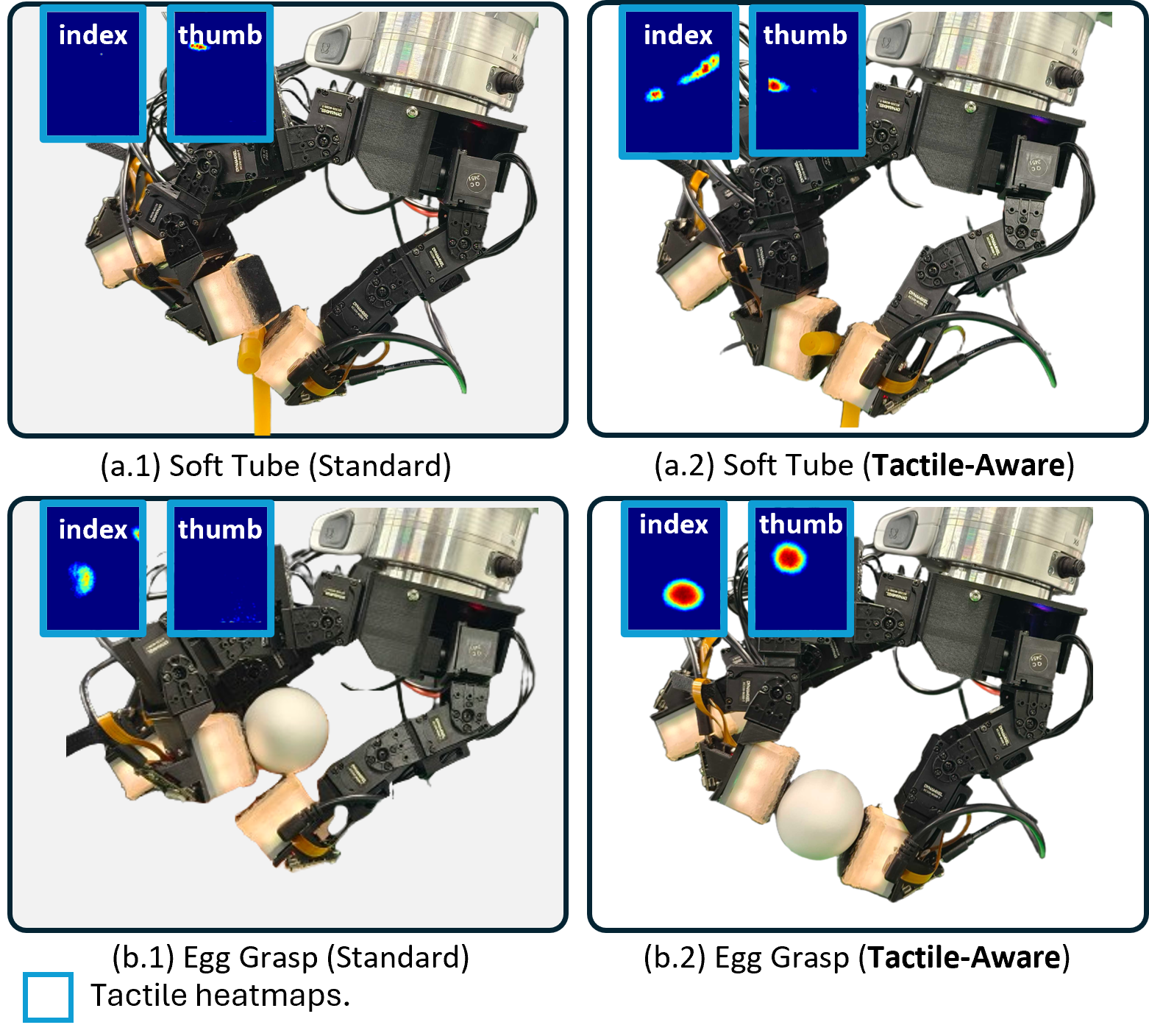}
\vspace{-4mm}
    \caption{Comparison of standard DexPilot retargeting vs. our tactile-aware extension on two manipulation tasks. (a) Soft tube grasping, (b) Egg grasping. Standard retargeting (left column) produces edge contact with negligible sensor signals, while our constraint (right column) enforces parallel fingertip alignment, enabling effective tactile sensing. Blue insets show tactile sensor heatmaps from thumb and index fingertips.}
    \label{fig:tactile_retarget}
\vspace{-3mm}
\end{figure}

The parallelism constraint can be toggled on/off before teleoperation based on task requirements—enabled for contact-critical tasks (pinching, rolling) and disabled for tasks requiring varied finger configurations (power grasps, hook grasps). As shown in Fig.~\ref{fig:tactile_retarget}, standard position-only retargeting produces edge contact that yields negligible tactile signals, while our tactile-aware constraint enforces parallel alignment that enables effective tactile sensing. Tactile sensor heatmaps (blue insets) confirm this improvement: standard retargeting shows minimal sensor activation, whereas our extension achieves robust contact engagement across both soft tube and egg manipulation tasks.

\subsection{Vibrotactile Haptic Feedback}
\label{sec:haptic_feedback}
The haptic feedback system converts tactile signals into vibrotactile cues through the following processing pipeline. For each fingertip $i$, let $\delta_i^{\text{max}}$ denote the maximum deformation depth measured by the 9DTact sensor array. The PWM command $u_i$ for the corresponding LRA is computed as:

\begin{equation}
u_i = \text{LPF}\left(\text{PWM}_{\max} \cdot \frac{\delta_i^{\text{max}} - \delta_{\min}}{\delta_{\max} - \delta_{\min}}\right)
\end{equation}

where $[\delta_{\min}, \delta_{\max}]$ are pre-calibrated deformation thresholds, and LPF denotes a first-order low-pass filter for signal smoothing. This monotonic mapping provides a readily perceivable contact-intensity cue; proportionality to force is only approximate under an elastic-contact assumption. PWM-LRA feedback is a practical low-cost choice, not an optimized encoding.

We mount LRAs on ring fixtures on the finger (rather than on the fingertip) to avoid interfering with fingertip contacts and vision-based tracking, while maintaining consistent placement and comfort. Despite the offset from the fingertip, users can localize finger-wise vibration due to spatial separation of stimulation sites.

We drive each LRA with a fixed carrier frequency set to the actuator’s resonance (from its datasheet) and modulate perceived intensity via PWM duty cycle. The parameters ($\text{PWM}_{\max}$ and depth thresholds) are chosen via short pilot trials to ensure perceptible but comfortable feedback and to avoid saturation during typical contacts.

\section{Experiments}

\begin{figure*}[t]
    \centering
    \includegraphics[width=\textwidth]{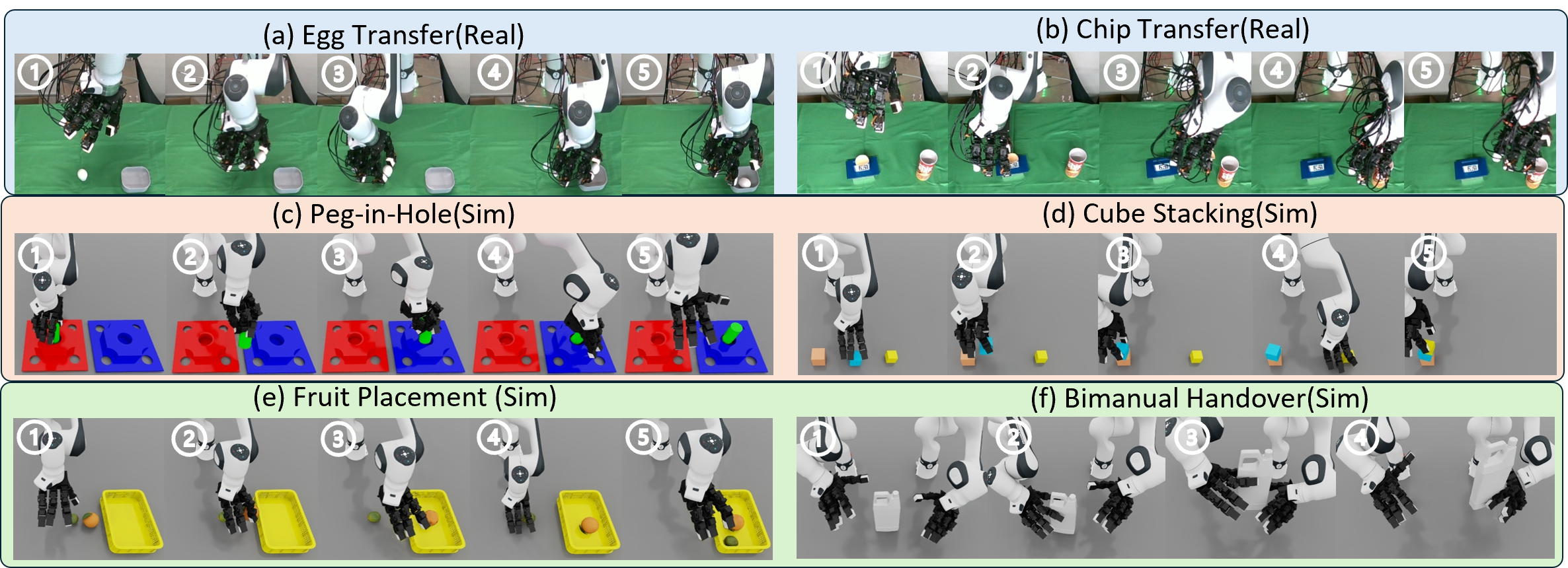}
    \caption{Experimental tasks for user study. Numbers indicate key stages in task progression. Real-world: (a) egg transfer to bowl, (b) fragile chip grasping. Simulation: (c) peg-in-hole insertion, (d) three-cube stacking, (e) fruit pick-and-place, (f) bimanual bottle handover.}
    \label{fig:tasks}
\vspace{-5mm}
\end{figure*}

\subsection{Evaluation Overview}
We evaluate ViHaTeleop through two complementary studies: (1) a controlled ablation comparing teleoperation performance with and without haptic feedback, and (2) imitation learning experiments assessing whether collected demonstrations support downstream policy training. Our experiments focus on contact-critical manipulation tasks.

\subsection{Implementation}
\textbf{Real Robot Setup.} Franka FR3 (7-DoF) equipped with LEAP Hand (16-DoF) and 9DTact tactile sensors on three fingertips (thumb, index and middle fingers). An Intel RealSense D455 RGB-D camera is mounted in front of the workspace. Control runs at 20 Hz. 

\textbf{Simulation Setup.} Isaac Sim 5.0 \cite{isaacsim} with both single-arm and dual-arm Franka FR3 configurations. A depth-based proxy approximates 9DTact tactile feedback for efficient simulation.

\subsection{User Study: Haptic Feedback Ablation}
\label{sec:user_study}

\textbf{Participants.} Nine volunteers (A–I) participated in the study. The sample included six males and three females, aged 22–31 years (mean $= 26.3$, SD $= 3.39$). Eight participants were right-handed. Regarding prior teleoperation experience, two reported extensive experience, one reported limited experience, and six reported no prior experience.  

We determined our sample size based on effect sizes observed in comparable haptic teleoperation studies. A priori power analysis (G*Power 3.1, Wilcoxon signed-rank test, two-tailed, $\alpha = 0.05$, power $= 0.80$) indicated that nine participants provide adequate power to detect large effects (Cohen's $d \geq 0.8$), which are typical in haptic feedback ablation studies where differences are perceptually salient.


\textbf{Tasks.} We selected tasks to cover common contact-critical subskills: fragile grasping (egg, chip), insertion/contact onset (peg-in-hole), multi-contact manipulation (stacking), and bimanual coordination (handover). Real-robot tasks are performed co-located with direct line-of-sight, while simulation uses a fixed monitor view to emulate camera-based demonstration collection in embodied-AI data factories.
Six contact-critical manipulation tasks requiring precise force modulation (Fig.~\ref{fig:tasks}):


\emph{Real world (operator beside the robot):} (a) \emph{Egg transfer}---pick up an egg and place it into a bowl; (b) \emph{Chip transfer}---pick up a fragile potato chip and place it into a tube container. \emph{Simulation (operator viewing the Isaac Sim GUI on a monitor):} (c) \emph{Peg-in-hole}---extract a tube from one hole and insert it into another; (d) \emph{Cube stacking}---stack three cubes into a tower; (e) \emph{Fruit pick-and-place}---place an orange and then a lime into the same basket; (f) \emph{Bimanual handover}---transfer a liquid container from the right hand to the left.

\textbf{Protocol.} After a 10-minute tutorial and practice session with ViHaTeleop on both real and simulated platforms, participants practiced each task until achieving at least one successful trial to avoid floor effects. They then performed 10 experimental trials per task: 5 without haptic feedback and 5 with haptic feedback, presented in randomized order to control for learning effects.


\textbf{Metrics.} \emph{Objective:} (1) success rate per condition (successful trials out of 5); (2) completion time over successful trials. \emph{Subjective:} post-condition 5-point Likert ratings (1=strongly disagree, 5=strongly agree) on \textit{ease of manipulation} (``It was easy to manipulate objects using the teleoperation system.''), \textit{contact clarity} (``I could clearly tell when the robot hand made contact with an object.''), and \textit{grasp confidence} (``I felt confident that the object was securely grasped in the robot hand.'').

\subsubsection*{Statistics}
Paired comparisons use the two-tailed Wilcoxon signed-rank test (significance at $p<0.05$). Per-task statistical power is limited at $n=9$; we report $p$-values for transparency rather than claiming per-task significance.

\begin{table}[t]
\centering
\caption{User Study: Success Rates}
\label{tab:success_rates}
\small
\begin{tabular}{lcccc}
\hline
\textbf{Task} & \textbf{w/o Haptic} & \textbf{w/ Haptic} & \textbf{$\Delta$} & \textbf{$p$} \\
\hline
\multicolumn{5}{c}{\textit{Real World}} \\
\hline
Egg transfer & 66.7\% & \textbf{80.0\%} & +13.3\% & .172 \\
Chip grasping & 46.7\% & \textbf{48.9\%} & +2.2\% & .516 \\
\hline
\multicolumn{5}{c}{\textit{Simulation}} \\
\hline
Peg-in-hole & 42.2\% & \textbf{51.1\%} & +8.9\% & .250 \\
Cube stacking & 24.4\% & \textbf{28.9\%} & +4.5\% & .500 \\
Fruit placement & 46.7\% & \textbf{57.8\%} & +11.1\% & .531 \\
Handover & 51.1\% & \textbf{66.7\%} & +15.6\% & .125 \\
\hline
\end{tabular}
\end{table}

\vspace{-7mm}
\begin{table}[t]
\centering
\caption{User Study: Completion Times (seconds)}
\label{tab:completion_times}
\small
\begin{tabular}{lcccc}
\hline
\textbf{Task} & \textbf{w/o Haptic} & \textbf{w/ Haptic} & \textbf{$\Delta$} & \textbf{$p$} \\
\hline
\multicolumn{5}{c}{\textit{Real World}} \\
\hline
Egg transfer & 27.1 & \textbf{23.6} & \mbox{--12.8\%} & .742 \\
Chip grasping & 26.9 & 28.5 & +5.9\% & 1.00 \\
\hline
\multicolumn{5}{c}{\textit{Simulation}} \\
\hline
Peg-in-hole & 35.5 & \textbf{31.3} & \mbox{--12.0\%} & .078 \\
Cube stacking & 55.0 & 62.6 & +13.8\% & .359 \\
Fruit placement & 37.9 & 39.6 & +4.4\% & .887 \\
Handover & 21.6 & 30.4 & +40.2\% & .297 \\
\hline
\end{tabular}
\vspace{-5mm}
\end{table}

\begin{table}[t]
\centering
\caption{Simulation Cube Success Rate Ablation: Default vs. Parallelism Constraint}
\label{tab:parallelism_ablation}
\small
\begin{tabular}{llccc}
\hline
\textbf{Haptic} & \textbf{Cube} & \textbf{Default} & \textbf{Parallelism} & \textbf{$\Delta$} \\
\hline
\multirow{2}{*}{w/o} & Blue & 35.0\% & \textbf{50.0\%} & +15.0\% \\
& Yellow & 5.0\% & \textbf{10.0\%} & +5.0\% \\
\hline
\multirow{2}{*}{w/} & Blue & 55.0\% & \textbf{75.0\%} & +20.0\% \\
& Yellow & 5.0\% & 5.0\% & 0.0\% \\
\hline
\end{tabular}
\vspace{-9.5mm}
\end{table}

\vspace{5mm}
\subsection{User Study Results}\label{subsec:userstudy}

\begin{figure*}[t]
    \centering
    \includegraphics[width=\textwidth]{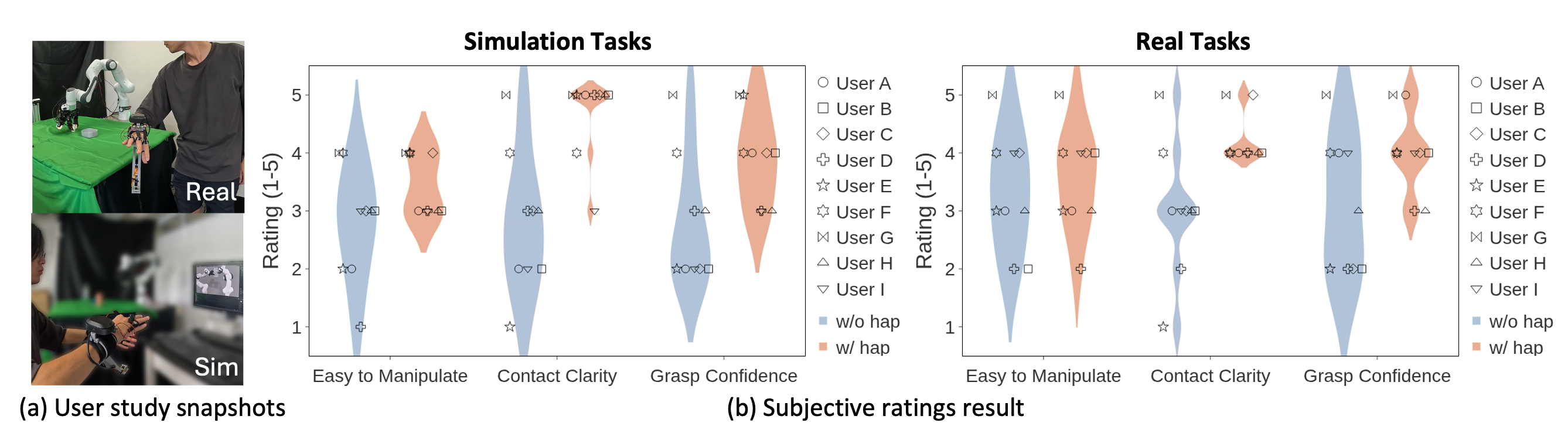}
\vspace{-8mm}
    \caption{User study experimental setup and subjective ratings. 
    (a) Experimental snapshots showing teleoperation in real-world (top) 
    and simulation (bottom) environments. (b) Subjective ratings comparing 
    without and with haptic feedback across nine participants. 
    Error bars indicate standard deviation. Statistical comparisons use two-tailed 
    Wilcoxon signed-rank tests; detailed values are reported in Sec.~\ref{subsec:userstudy}.}
    \label{fig:user_study}
\vspace{-6mm}
\end{figure*}
\textbf{Success Rates.} Table~\ref{tab:success_rates} shows descriptive success-rate trends across all tasks. In this 9-participant dataset, haptic feedback increased success rates on all six tasks, with gains ranging from +2.2 to +15.6 percentage points. The largest gains were observed in simulation handover (+15.6 points) and real-world egg transfer (+13.3 points), while chip grasping showed a small gain (+2.2 points).

\textbf{Completion Times.} Table~\ref{tab:completion_times} shows task-dependent effects on completion time. Completion time decreased for egg transfer (27.1s $\to$ 23.6s, --12.8\%) and peg-in-hole (35.5s $\to$ 31.3s, --12.0\%), but increased for chip grasping (+5.9\%), cube stacking (+13.8\%), fruit placement (+4.4\%), and handover (+40.2\%). Notably, single-contact tasks (egg transfer, peg-in-hole) showed both faster completion and higher success, while multi-stage tasks (cube stacking, fruit placement, handover) showed higher success at the cost of increased time. We analyze this task-dependent pattern in Sec.~\ref{sec:discussion}.

\textbf{Parallelism-Constraint Ablation.} Table~\ref{tab:parallelism_ablation} reports a focused ablation on the simulation cube task using a subset of four participants (F–I). We compare default retargeting and retargeting with the fingertip parallelism constraint under otherwise identical conditions. The parallelism constraint improved blue-cube success in both conditions (w/o haptic: 35.0\% $\to$ 50.0\%, +15.0 points; w/ haptic: 55.0\% $\to$ 75.0\%, +20.0 points). For yellow-cube attempts, performance improved from 5.0\% to 10.0\% without haptics and remained unchanged at 5.0\% with haptics. These results are consistent with the intended benefit of parallel fingertip contact alignment for contact-critical grasp stages.

\textbf{Subjective Experience.} Figure~\ref{fig:user_study}b shows subjective ratings on 5-point Likert scales (1=strongly disagree, 5=strongly agree) across three dimensions. In simulation, ease improved from 2.78$\pm$0.97 to 3.44$\pm$0.53 ($p<0.05$), contact clarity improved from 2.78$\pm$1.20 to 4.67$\pm$0.71 ($p=0.0117$), and grasp confidence improved from 2.78$\pm$1.09 to 3.89$\pm$0.78 ($p=0.0278$). In real-world operation, ease changed from 3.33$\pm$1.00 to 3.56$\pm$0.88 ($p=0.3173$, n.s.), contact clarity improved from 3.00$\pm$1.12 to 4.22$\pm$0.44 ($p=0.0112$), and grasp confidence improved from 3.11$\pm$1.17 to 4.00$\pm$0.71 ($p=0.0276$). Overall, the subjective results indicate consistent gains in contact perception and confidence.

\subsection{Imitation Learning Validation}
\label{sec:imitation}
\begin{figure}[t]
    \centering
    \includegraphics[width=\columnwidth]{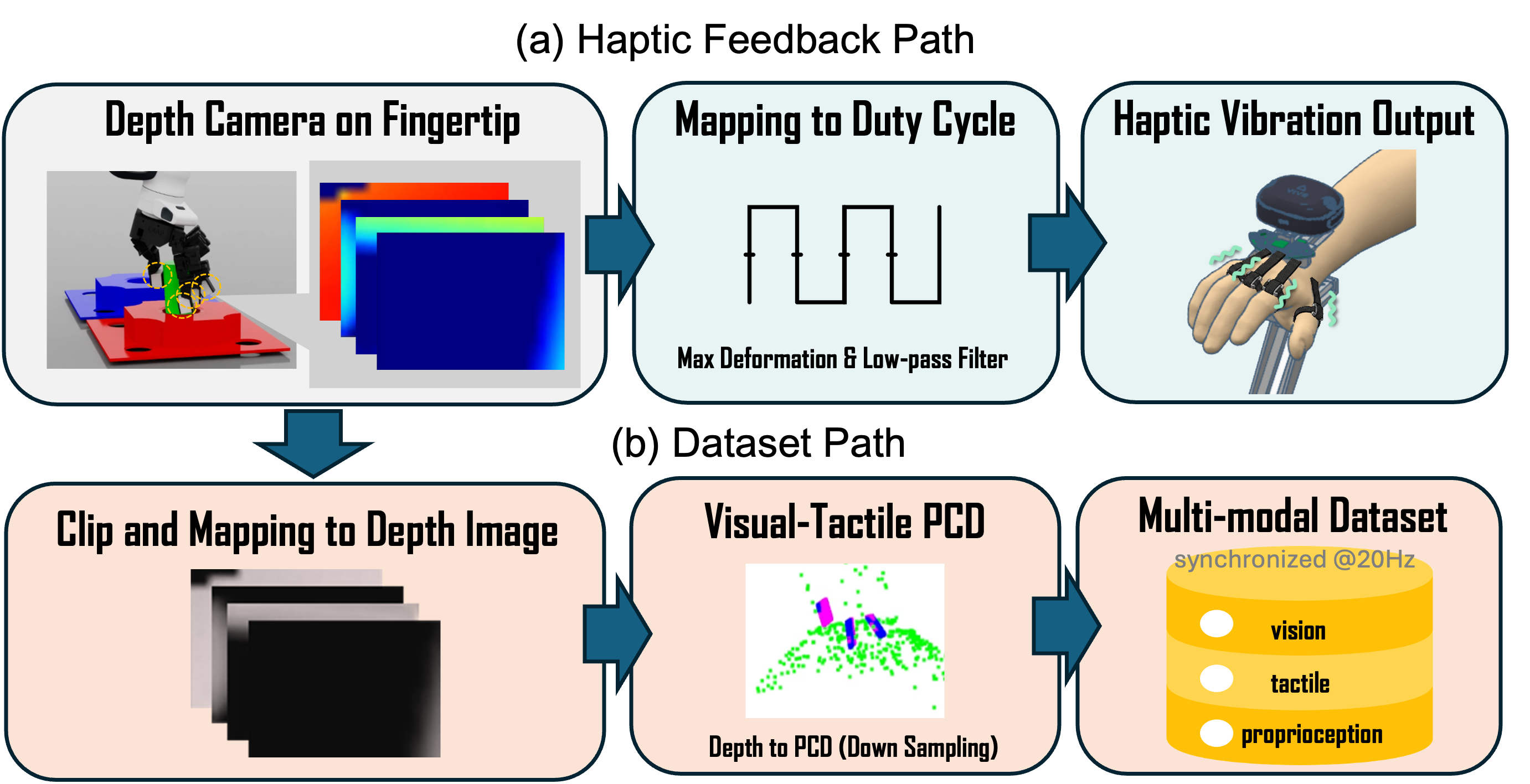}
    \caption{Dual-purpose tactile proxy pipeline in 
    simulation. A depth camera on each fingertip 
    approximates 9DTact deformation. Top path (blue): 
    the signal drives real-time haptic feedback to 
    the operator via max-deformation extraction and 
    PWM mapping. Bottom path (beige): the same 
    depth data is converted to tactile point clouds 
    and fused with visual observations into a 
    synchronized multi-modal demonstration dataset 
    for downstream policy learning.}
    \label{fig:tactile_proxy}
\vspace{-6mm}
\end{figure}

As shown in Fig.~\ref{fig:tactile_proxy}, the same depth-camera proxy that drives operator haptic feedback also provides tactile point clouds for policy learning, enabling a unified pipeline from data collection to training. We extend 3D-ViTac's \cite{3dvitac} visual-tactile representation to dexterous hands with multi-fingertip tactile sensing. The unified point cloud representation integrates visual and tactile information: each point contains $(x, y, z)$ spatial coordinates, a one-hot encoding indicating point type (visual or tactile), and normalized force/deformation depth from tactile sensors. For dexterous hands, tactile point clouds from three fingertips are transformed to the global frame via forward kinematics and concatenated with visual point clouds from RGB-D cameras (Fig.~\ref{fig:tactile_proxy}, bottom path). The rest of the policy architecture (encoder and diffusion decoder) remains the same as 3D-ViTac. 

\textbf{Training Data.} We collected approximately 50 demonstrations from three tasks: (c) peg-in-hole, (d) cube stacking, and (e) fruit pick-and-place using ViHaTeleop. Each demonstration includes synchronized RGB-D images, 9DTact tactile data, and robot states and actions all recorded at 20 Hz. 

\textbf{Training Details.} All policies were trained for 3,000 epochs. We report results using the final checkpoint for all methods to ensure fair comparison.

\textbf{Baselines.} We compare two conditions: (1) vision-only and (2) visual-tactile.

\begin{table}[t]
\centering
\caption{Imitation Learning Policy Rollout Results}
\label{tab:policy_rollout}
\small
\begin{tabular}{lccc}
\hline
\textbf{Task} & \textbf{Vision} & \textbf{Visual-Tactile} & \textbf{$\Delta$} \\
\hline

Peg-in-hole & 10\%& \textbf{27\%}& \textbf{+17\%}\\
Cube stacking & 27\%& 36\%& +9\%\\
Fruit placement & 11\%& 9\%& -2\%\\
\hline
\textbf{Average} & \textbf{16\%} & \textbf{24\%} & \textbf{+8\%} \\
\hline
\multicolumn{4}{l}{\scriptsize Results from 100 rollout episodes per task.} \\
\end{tabular}
\vspace{-8mm}
\end{table}

\textbf{Results.} Table~\ref{tab:policy_rollout} shows rollout success rates across 100 evaluation episodes per task. Visual-tactile policies achieved 24\% average success compared to 16\% for vision-only (+8 percentage points), with task-dependent behavior. The largest improvement appears on peg-in-hole (10\% $\to$ 27\%, +17 points), cube stacking also improved (27\% $\to$ 36\%, +9 points), and fruit placement decreased slightly (11\% $\to$ 9\%, --2 points). These results suggest tactile cues are most helpful for contact-critical subtasks in this dataset, while additional tactile input may provide limited benefit for more vision-dominant tasks under the current data scale.


\subsection{Discussion}\label{sec:discussion}

The updated 9-participant study shows mixed objective outcomes and consistent subjective gains. Success rate increased across all six tasks, but completion time improved in only two tasks and increased in four. By contrast, subjective ratings showed statistically significant improvements in contact clarity and grasp confidence in both simulation and real-world settings. Within our tested protocol, these results support that low-cost vibrotactile feedback ($<$\$10 per hand) can improve teleoperation quality without motorized force-feedback hardware. The contrast between non-significant per-task objective tests and significant subjective results is consistent with statistical power: per-task 
comparisons use 5 trials per condition per participant, while subjective ratings aggregate each participant's holistic experience across all trials.

\textbf{Visual-Haptic Sensory Trade-offs.} A consistent pattern in subjective data is that gains were generally larger in simulation than in real-world operation for contact clarity (+1.89 vs. +1.22) and grasp confidence (+1.11 vs. +0.89), while ease changes were smaller and not statistically significant in either setting. A plausible interpretation is that the simulation interface imposed stronger visual constraints, increasing the marginal utility of tactile cues. 

This suggests a practical design direction: in co-located settings with richer visual cues, haptics may play a complementary role; in constrained-view settings, haptics may provide larger relative gains. Dedicated experiments that explicitly vary visual conditions are needed before making stronger substitution claims.

\textbf{Speed-Accuracy Tradeoff and Task Structure.} The completion-time results are consistent with the speed-accuracy tradeoff reported in haptic teleoperation~\cite{yip_haptic_tradeoff}, where richer sensory feedback encourages more deliberate manipulation strategies. However, the tradeoff is not uniform: single-contact tasks (egg transfer, peg-in-hole) showed \emph{both} faster completion and higher success, while multi-stage tasks (cube stacking, fruit placement, handover) showed higher success at the cost of increased time. A plausible explanation is that for single-contact tasks, haptic feedback reduces exploratory retries by confirming contact onset on the first attempt, whereas for multi-stage tasks, operators process haptic cues at each contact transition, accumulating a time cost across stages even as each stage becomes more reliable. We also note that completion times are computed only over successful trials, so the non-haptic baseline selectively reflects fast attempts that were actually successful—the numerous quick failures are left out of the time comparison.

\textbf{Imitation Learning Validation.} The policy experiments provide preliminary evidence that ViHaTeleop demonstrations can support downstream multimodal learning. The largest observed gain was on peg-in-hole (+17 points over vision-only), indicating that tactile information can improve contact-critical tasks in our setup.

The task-dependent patterns reflect modality trade-offs: tactile benefits contact-critical subtasks (insertion, stacking) but, with limited training data, may not improve vision-driven tasks (fruit localization). These results provide preliminary support for the pipeline from haptic teleoperation to multimodal policy learning, while highlighting open questions about selective modality fusion and data-efficient multimodal learning. The simulation environments, tactile proxy, and demonstration datasets will be released to support future research in visual-tactile dexterous manipulation.

\textbf{Limitations.} We evaluate through controlled ablations within our system architecture rather than cross-system comparisons, which would introduce multiple confounding variables. Future work could explore comparative benchmarking across different teleoperation paradigms and systematically vary visual feedback quality to further characterize the visual-haptic trade-off. Remaining limitations include no external ground-truth benchmark of tracker or camera accuracy, single-participant LED evaluation, a small parallelism-constraint ablation, and a limited user-study sample ($n=9$).

\vspace{-2mm}
\section{CONCLUSIONS}

We presented ViHaTeleop, a low-cost (\$550), lightweight (0.7 kg) teleoperation system that integrates haptic feedback into a vision-based platform optimized for contact-critical manipulation data collection.

Our user study (Sec.~\ref{sec:user_study}) with nine participants across six contact-critical tasks showed that haptic feedback increased success rates across all tasks (+2.2 to +15.6 points), while completion-time effects were task-dependent (Tables~\ref{tab:success_rates}, \ref{tab:completion_times}). Subjective ratings (Fig.~\ref{fig:user_study}) showed significant improvements in contact clarity and grasp confidence in both simulation and real-world settings, with non-significant changes in ease.

The haptic integration pipeline—consisting of tactile-aware retargeting 
(Fig.~\ref{fig:tactile_retarget}), multi-fingertip 9DTact sensing, and LRA feedback mapping (Sec.~\ref{sec:haptic_feedback})—combined with targeted optimizations (fisheye camera for reduced torque, LED lighting for illumination robustness) achieves a favorable balance across cost, weight, and performance for large-scale data collection.

Our imitation learning experiments (Sec.~\ref{sec:imitation}) provided preliminary pipeline validation from teleoperation data collection to policy execution. Visual-tactile policies trained on ViHaTeleop demonstrations achieved 24\% average success versus 16\% for vision-only (Table~\ref{tab:policy_rollout}), indicating potential benefits of tactile information for contact-critical tasks in this dataset.

The simulation-integrated pipeline---including task environments, a depth-camera tactile proxy, and multi-fingertip point cloud registration---provides reusable infrastructure for future visual-tactile research with dexterous hands.

The implementation will be open-sourced. Future work will scale to larger datasets and diverse robot morphologies.


\section*{ACKNOWLEDGMENT}
This work was partially supported by JST, the establishment of university fellowships towards the creation of science technology innovation, Grant Number JPMJFS2102. The authors thank Daimon Robotics for their assistance with the design of the LRA driver board.

\end{document}